\documentclass{article}

\usepackage{arxiv}

\usepackage[utf8]{inputenc} 
\usepackage[T1]{fontenc}    
\usepackage{hyperref}       
\usepackage{url}            
\usepackage{booktabs}       
\usepackage{amsfonts}       
\usepackage{nicefrac}       
\usepackage{microtype}      

\usepackage{graphicx}
\usepackage{natbib}
\usepackage{doi}

\usepackage{amsmath,amssymb} 
\usepackage{diagbox,multirow} 
\usepackage{xcolor}
\usepackage{siunitx}
\usepackage[font=it]{caption}
\title{Patient-specific vs Multi-Patient Vision Transformer for Markerless Tumor Motion Forecasting}

\author{ \href{https://orcid.org/0009-0003-3753-177X}{\includegraphics[scale=0.06]{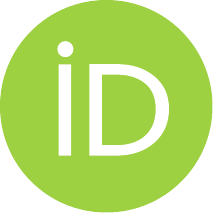}\hspace{1mm}Gauthier Rotsart de Hertaing} \\
	ICTEAM\\
	UCLouvain\\
	Louvain-la-Neuve, 1348, Belgium \\
	\texttt{gauthier.rotsart@uclouvain.be} \\
	\And
	\href{https://orcid.org/0000-0001-9034-0794}{\includegraphics[scale=0.06]{orcid.pdf}\hspace{1mm}Dani Manjah} \\
	ICTEAM\\
	UCLouvain\\
	Louvain-la-Neuve, 1348, Belgium \\
	\texttt{dani.manjah@uclouvain.be} \\
    \And
	\href{https://orcid.org/0000-0002-7243-4778}{\includegraphics[scale=0.06]{orcid.pdf}\hspace{1mm}Benoit Macq} \\
	ICTEAM\\
	UCLouvain\\
	Louvain-la-Neuve, 1348, Belgium \\
	\texttt{benoit.macq@uclouvain.be}
}

\renewcommand{\shorttitle}{\textit{arXiv} Template}

\hypersetup{
pdftitle={A template for the arxiv style},
pdfsubject={q-bio.NC, q-bio.QM},
pdfauthor={David S.~Hippocampus, Elias D.~Striatum},
pdfkeywords={First keyword, Second keyword, More},
}

\begin{document}
\maketitle

\begin{abstract}
	\noindent {\bf Background:} Accurate forecasting of lung tumor motion is essential for precise dose delivery in proton therapy. While current state-of-the-art markerless methods mostly rely on deep learning models, the use of transformer-based architectures remains unexplored in this domain, despite their proven ability in trajectory forecasting tasks. \\
{\bf Purpose:} This work introduces a novel markerless forecasting approach for lung tumor motion using Vision Transformers (ViT). Two training strategies are compared under clinically realistic constraints: a patient-specific (PS) approach that learns individualized motion patterns, and a multi-patient (MP) model designed for generalization. The comparison explicitly accounts for the limited number of images that can be generated between planning and treatment sessions, a key constraint in clinical implementation.
 \\
{\bf Methods:} Digitally reconstructed radiographs (DRR) derived from planning four-dimensional computed tomography scans (4DCT) of 31 patients were used to train the MP model; a 32nd patient was reserved for evaluation. PS models were trained using only the target patient’s planning data. Both models used 16 DRRs per input and predicted tumor motion over a 1-second horizon. Performance was assessed using Average Displacement Error (ADE) and Final Displacement Error (FDE), on both planning (T1) and treatment (T2) data. \\
{\bf Results:} On T1 data, PS models outperformed MP models across all training set sizes, especially when trained on larger datasets (up to 25,000 DRRs, $p < 0.05$). However, MP models demonstrated stronger robustness to inter-fractional anatomical variability and achieved comparable performance on T2 data, without requiring retraining.  \\
{\bf Conclusions:} To the best of our knowledge, this study is the first to apply ViT architectures to markerless tumor motion forecasting. It demonstrates that while patient-specific training can achieve higher precision, multi-patient models offer robust out-of-the-box performance, which is particularly suitable for time-constrained clinical settings. Future work may investigate fine-tuning MP models on each new patient, with the goal of enabling rapid and accurate model deployment.
\end{abstract}


\section{Introduction}
Proton therapy aims to target cancerous cells with protons while minimizing exposure to surrounding healthy tissues and critical structures such as the heart and spine.  Its success depends on locating tumors precisely, which is particularly challenging for thoracic and abdominal tumors, which move with the patient's breathing. This movement introduces uncertainties in tumor positioning, compelling oncologists to apply safety margins to ensure proper dose delivery and \textbf{leading in turn to the exposure of healthy tissues and vital structures}\cite{dieterich2007tumor}. Various techniques have been explored to improve tumor localization and minimize damage to healthy tissues, including breath-holding and abdominal compression \cite{murphy2002effectiveness, nelson2005respiration, nissen2013improved, lin2017evaluation}, as well as respiratory gating \cite{shirato2000physical, berbeco2005residual, keall2006management}. Nevertheless, breath-holding is not feasible for all patients, and respiratory gating tends to increase treatment time. To overcome these limitations, this paper focuses on a novel real-time tumor tracking (RTTT) method that aims to enable dynamic beam adjustments to the tumor's position during treatment.

RTTT methods rely on extracting tumor position from X-ray fluoroscopy imaging using either marker-based or markerless approaches. However, due to the low soft-tissue contrast of fluoroscopy images, marker-based tracking methods are often employed to enhance localization accuracy\cite{shirato2003feasibility, dieterich2007tumor}. These methods have been extensively researched and are known for delivering state-of-the-art results in tumor motion forecasting\cite{putra2006prediction, johl2020performance, krauss2011comparative, jiang2019prediction, teo2018feasibility, wang2018feasibility, lin2019towards, pohl2022prediction}. However, they are invasive, requiring surgical procedures that can lead to complications such as pneumothorax and infections \cite{nuyttens2006lung, prevost2008endovascular, abdefg2002ct, li2009feasibility, prevost2009stereotactic}. They also rely on real-time extraction of marker positions and perfect correlation with the tumor centroid, which requires constant updating due to inter-fraction changes. Conversely, markerless methods based on fluoroscopic images are being explored, offering new perspectives for tumor motion management in proton therapy.  



One approach to markerless tracking leverages the correlation between the tumor and the diaphragm \cite{cervino2009diaphragm, cervino2010tumor, petkov2013robust, petkov2014diaphragm, hirai2020regression}. Another involves template matching between fluoroscopic images and four-dimensional computed tomography scans (4DCT) \cite{li2009feasibility, de2021markerless, de2021markerless2, fu2024deep}. Recent advancements in deep learning have shown promising results in providing information about tumor positions that are typically invisible without fiducials. Despite these advancements, addressing anatomical changes between treatment sessions remains a key focus \cite{hirai2019real, sakata2020machine, zhao2019markerless, terunuma2023explainability, wei2019convolutional, wei2020real}. Moreover, the aforementioned markerless methods neglected system latency due to their primary focus on extracting tumor positions from fluoroscopic images. Nevertheless, latency can reach up to 500 ms\cite{poulsen2010detailed}, which impacts dosimetry and has been shown to reduce clinical target volume coverage\cite{bedford2015effect}.

As the field of deep learning continues to evolve to address the challenges of tracking mobile tumors, innovative architectures are being explored. Among them, transformers — a particular class of deep neural networks \cite{vaswani2017attention, dosovitskiy2020image} — are emerging as promising tools in tasks that involve sequential data, especially in trajectory forecasting \cite{giuliari2021transformer, franco2023under, rotsart2024trajvivit}. To the best of our knowledge, this work is the first to investigate the ability of transformer networks to track and forecast the trajectories of mobile tumors. 

We contribute a novel end-to-end markerless method that focuses on forecasting tumor motion. The model is trained on digitally reconstructed radiograph (DRR) images, simulating fluoroscopic image acquisition with a realistic frame rate. To ensure clinical relevance, the model is trained on images from the treatment planning phase and evaluated on images acquired later in the treatment process. In addition, we explore training strategies, comparing patient-specific and multi-patient approaches, considering the limited time available between the planning and treatment sessions. Our findings showed that \textbf{our multi-patient transformer network provides an accurate markerless method that is robust to system latency and suitable for clinical settings}. Additionally, in line with fostering medical research reproducibility, we provide a GitHub repository available at \\ \href{https://github.com/GauthierRotsart/Tracking\_tumor}{https://github.com/GauthierRotsart/Tracking\_tumor}, advancing markerless tracking methods in clinical settings. 



\section{Materials and Methods}
Sections \ref{sec:FDG_study} and \ref{sec:CPAP_study} describe the datasets used in this work. They include the patients, the treatment analyzed and the breathing amplitude computed in the $x,y,$ and $z$ directions with Euclidian norm and the volume of the tumor. The generation of the DRRs is explained in Section \ref{sec:dataset_creation}.
\subsection{Patients selection}
Data from thirty-two patients with lung tumors were collected in two studies \cite{di2017correlation,di2018effect}. Each patient's thoracic region was scanned using 4DCT with a slice thickness of 2mm under free-breathing conditions, and the scans were divided into ten equally distributed temporal phases. Audio coaching was used to ensure regular breathing. Subsequently, a time-averaged mid-position (MidP) CT was generated using the OpenTPS \cite{wuyckens2023opentps} software to characterize tumor motion associated with respiration \cite{janssens2011diffeomorphic}. This image and the associated deformation fields form a patient-specific motion model. An experienced oncologist delineated the gross tumor volume (GTV). The center of mass of the GTV was used as the tumor position in the subsequent analysis.
\subsubsection{Small tumor motion dataset (SM dataset)}\label{sec:FDG_study}
Fourteen patients with unresectable lung cancer underwent 4DCT scans at two different time points \cite{di2017correlation}. The first scan ($T1$) was conducted during the planning session, with a median of ten days before the start of radiotherapy. The subsequent scan took place during the second week ($T2$) of treatment. In this dataset, the tumors vary in size and are predominantly located in the upper lung, resulting in relatively small respiration-induced tumor motion\cite{liu2007assessing}. The median breathing amplitude is 2.98 mm for $T1$ and 2.44 mm for $T2$, with a narrow range. In contrast, the gross tumor volume shows a wide range, varying from 13 mL to 280 mL in $T1$ and from 8 mL to 200 mL in $T2$.
\subsubsection{Small tumor volume dataset (SV dataset)}\label{sec:CPAP_study}
Eighteen patients with lung tumors or lung metastases underwent 4DCT scans at two different time points: once during the planning session ($T1$) and again on the first day of treatment ($T2$), with a median interval of ten days between the two sessions \cite{di2018effect}. This dataset features tumors with relatively small gross tumor volumes, with medians of 4.19 mL for $T1$ and 5.77 mL for $T2$. However, the breathing amplitude varies significantly, ranging from 4 to 26 mm in $T1$ and from 3 to 35 mm in $T2$.
\subsection{Methodology}
Our markerless forecasting method involves training a vision transformer network. We investigated two training strategies for this network: a patient-specific approach and a multi-patient approach, aiming to determine whether it was more beneficial for the model to learn patient-specific characteristics or to generalize across multiple patients. To ensure a fair comparison, both strategies were trained with the same number of data samples (i.e., constant iteration count), and evaluated on planning (T1) and treatment (T2) sessions. To create the training dataset, the planning 4DCT of each patient was processed using the public software OpenTPS \cite{wuyckens2023opentps}, generating corresponding fluoroscopy sequences. Subsequently, an adaptation of the vanilla vision transformer model \cite{rotsart2024trajvivit} was trained for 100 epochs under both strategies.
\subsubsection{Data processing} \label{sec:dataset_creation}
\subsubsection*{Training dataset}

Using the averaged motion PS model derived from a 4DCT, a patient-specific training dataset was created with tools from OpenTPS. For each patient anatomy, this software generates a synthetic breathing signal consisting of $\alpha$ points, which is then applied to simulate motion. This signal is applied at the target center and combined with deformation fields to simulate overall respiratory motion. As a result, a sequence of $\alpha$ synthetic intra-fraction 3DCT images is generated. The DRRs are then computed through projection onto the coronal axis. The Beer-Lambert absorption-only model is used to simulate realistic fluoroscopy images. 

\begin{figure}[!ht]
    \centering    
    \includegraphics[width=\linewidth]{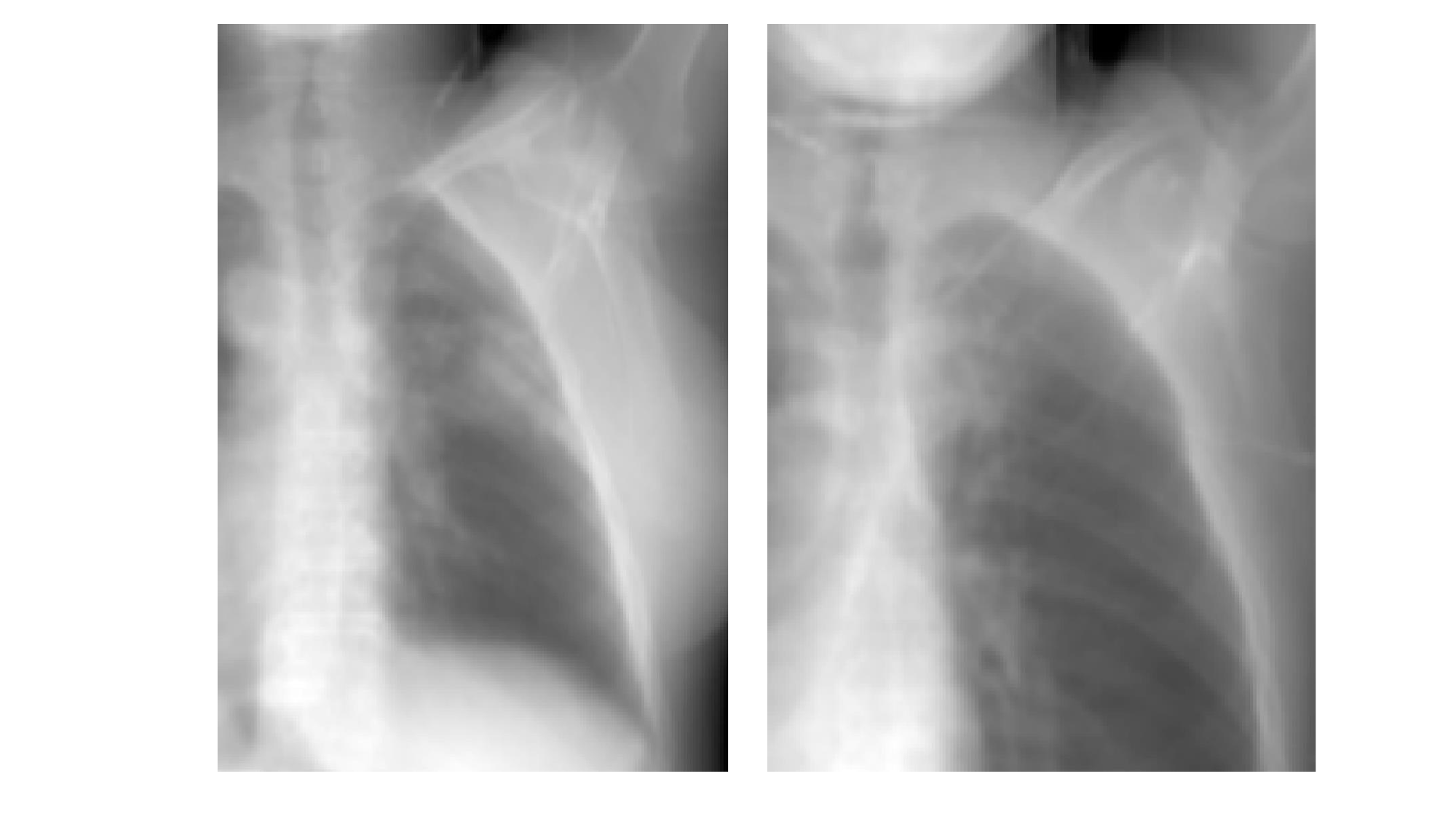}
    \caption{Two cropped DRRs are illustrated. The safety margins ensure the preservation of most of the tissues in the tumor-affected lung.}
    \label{fig:XF_images}
\end{figure}

The DRRs were cropped around the target contour in the midP image with arbitrary margins of 50 mm on the axial axis and 100 mm on the sagittal axis, optimizing the data generation process by reducing computational time while preserving most of the lung tissue that contains the tumor. The DRRs were then normalized to values between $0$ and $1$. The tumor's position was centered around zero, using the center of mass of the gross tumor volume in the planning midP image as the reference position. Each coordinate ($x$, $y$, $z$) was then normalized by its corresponding motion amplitude ($A_x$, $A_y$, $A_z$), derived from the planning 4DCT. This normalization serves two main purposes: improving training convergence by ensuring consistent scaling of all input features, and mitigating the impact of significant differences in motion amplitudes, which are often observed between $A_z$ and $A_x$, $A_y$. Examples of trained images coming from two patients are illustrated in Figure \ref{fig:XF_images}. 

\subsubsection*{Test dataset}
Testing was performed on 10 free-breathing sequences (20s each) from the same patient. These sequences were co-registered with the reference set (i.e., the planning session) by applying rigid registration between the MidP images of each 4DCT, simulating realistic patient setup errors. The test set images were cropped using the same bounding box as in the training set. Due to baseline shifts and setup variability, the tumor may not be centered in the cropped region. The center of mass of the GTV was centered and normalized relative to the tumor's position in the midP image.
\subsubsection{Model architecture and training}
The following section details the model and its training procedure. The evaluation metric used to assess performance is introduced in Section~\ref{sec:metric}.
\subsubsection*{Model architecture}
Vanilla Vision Transformers \cite{dosovitskiy2020image} are typically employed for classification tasks and consist of a single encoder and a head layer. In this work we used the TrajViViT network \cite{rotsart2024trajvivit}, which replaces the head layer with a decoder for regression tasks. The encoder's role is to embed the input images into a feature space, while the decoder autoregressively predicts the motion. Both components comprise six layers, similar to vanilla transformers \cite{vaswani2017attention}. Typically, the embedding size is $512$, and eight attention heads are used. The inner-layer of the feed forward network is composed of $2048$ neurons. The architecture of the network is depicted in Figure \ref{fig:trajvivit}: 
\begin{figure}[htpb]
    \centering    
    \includegraphics[width=\linewidth]{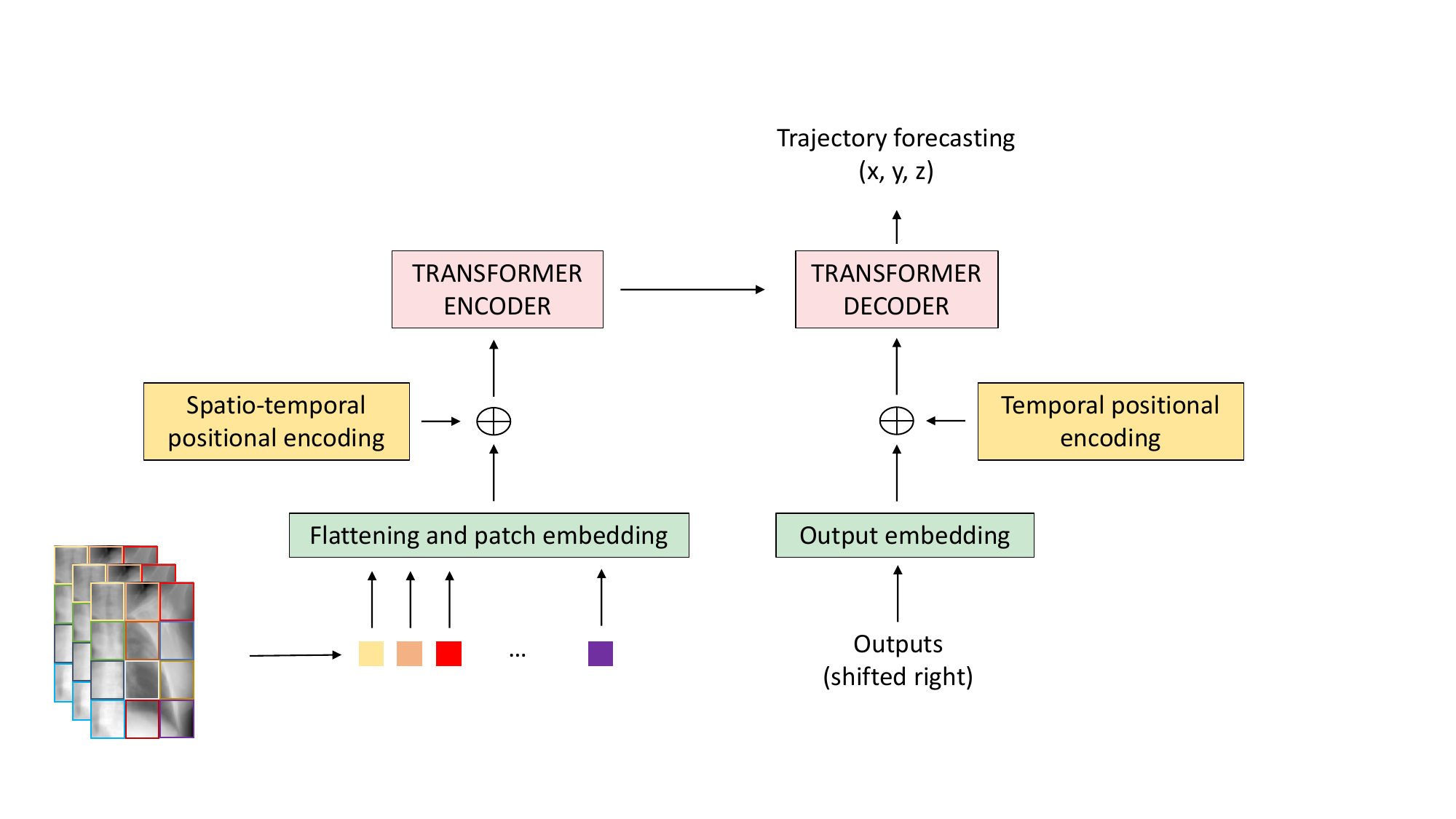}
    \caption{The architecture of TrajViViT is illustrated. First, the sequence of fluoroscopy images is divided into non-overlapping patches. Then, a spatio-temporal positional encoding is applied to encode the position of each patch both spatially (for the image) and temporally (for the sequence). These encoded patches are processed by a vanilla encoder-decoder transformer to predict the subsequent positions in an autoregressive manner.}
    \label{fig:trajvivit}
\end{figure}
\subsubsection*{Model optimization}
Transformers were trained on $100$ epochs with a batch size of $16$, using time series of $16$ past images as input. A cosine annealing scheduler \cite{loshchilov2016sgdr} was used, with a warmup period of ten epochs during which the learning rate increased linearly from $\num{5e-7}$ to $\num{5e-5}$. After the warmup, the learning rate decreased back to $\num{5e-7}$ without restarts. The loss function was root mean square error (RMSE) averaged over all forecasted time-points. The Adam optimizer was employed, with Glorot initialization\cite{glorot2010understanding}, and a dropout rate of 0.1. Teacher forcing \cite{williams1989learning} was utilized to expedite training, although this approach can cause some leakage during inference as the transformer encoder receives future information when updating the weights. 
\subsubsection*{Training strategies}
This study investigate two training strategies for forecasting tumor positions: a patient-specific approach and a multi-patient approach. The patient-specific strategy aims to learn individualized features, such as motion amplitude or the contrast between lung and tumor tissue. However, one of its limitations is the restricted training window between planning and treatment sessions, which may hinder model optimization. In contrast, the multi-patient strategy leverages a broader data distribution across multiple individuals and can be trained offline, making it less sensitive to time constraints. Nevertheless, capturing inter-patient variability can be challenging and may reduce precision for specific cases. \\

To ensure a fair comparison between the strategies, the models were trained with the same number of iterations. The multi-patient models were trained using Leave-One-Out Cross-Validation. All models were trained from scratch and evaluated across multiple random seeds to ensure robust results.
\subsubsection{Forecasting the tumor position}\label{sec:metric}
The delay between image acquisition and beam delivery necessitates forecasting tumor positions with a prediction horizon that exceeds system latency. This ensures that the forecast positions remain accurate despite the inherent time gap. In proton therapy, where treatment involves variable operations such as beam energy modulation or dose delivery, predicting multiple future time points becomes particularly advantageous. This enables adaptation to the different durations required by each action and improves tumor tracking precision. Accordingly, performance is reported as the average displacement error (ADE) over a prediction horizon $\Delta$, corresponding to the prediction of $T_{pred}$ successive time points.

\begin{align}
    \text{ADE} &= \frac{\sum_{i=1}^{T_{pred}}\sqrt{(x_i-\hat{x}_i)^2+(y_i-\hat{y}_i)^2+(z_i-\hat{z}_i)^2}}{T_{pred}} \\
    \text{FDE} &= \sqrt{(x_{T_{pred}}-\hat{x}_{T_{pred}})^2+(y_{T_{pred}}-\hat{y}_{T_{pred}})^2+(z_{T_{pred}}-\hat{z}_{T_{pred}})^2}
\end{align}

where $(\hat{x}_i,\hat{y}_i,\hat{z_i})$ are the predicted position in millimeters at the horizon $i$ and $(x_i,y_i,z_i)$ the respective ground truth position. In addition to ADE, we also report the Final Displacement Error (FDE), defined as the Euclidean distance between the predicted and ground truth positions at the final time step of the prediction horizon.

\section{Results}
In the following section, we report the forecasting performance of patient-specific models, each trained on DRRs from the T1 distribution of a given patient $p$. These are compared with multi-patient models. Evaluation is conducted over five different random seeds to account for variability due to weight initialization. All results are reported with a prediction horizon of one second ($\Delta = 1$), corresponding to $T_{pred}$ = 5 future time-points acquired at 5 frames per second.
\subsection{Training data quantity}
To evaluate the performance of patient-specific (PS) and multi-patient (MP) models, we investigated how model accuracy varied as a function of the training data quantity. PS models are trained individually on data from the target patient, allowing them to specialize to that patient's anatomy. MP models, by contrast, are trained on data from multiple other patients, and evaluated on the target patient without having seen any of their data during training. 

For both approaches, we varied the amount of training data — from 1,000 to 25,000 DRRs — and report the performance averaged across all target patients. The upper limit of 25,000 DRRs corresponds to the maximum amount of training data that can be generated and utilized before the first treatment session, simulating the available time (typically one day) between the planning and the first treatment session. This time constraint reflects a realistic clinical scenario where models must be trained prior to treatment initiation. This setup enables a direct comparison between patient-specific specialization and multi-patient generalization under equal data constraints. 

\begin{figure}[htpb]
    \centering    
    \includegraphics[width=\linewidth]{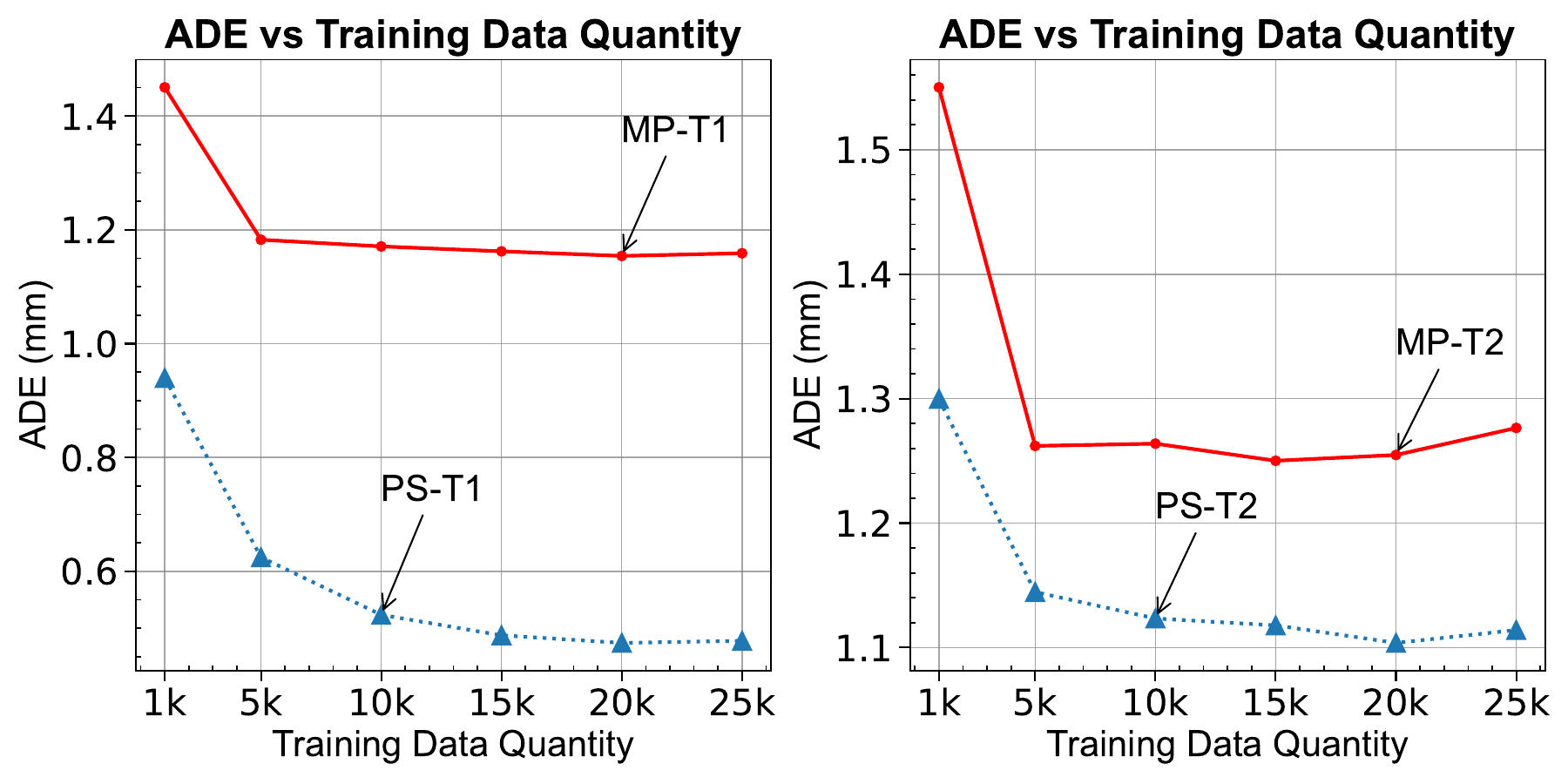}
    \caption{Comparison of patient-specific and multi-patient training in terms of training data quantity: PS models outperform MP models during the planning session (T1). However, MP models indicate more robustness to inter-fractional variations. The performance is evaluated using the average displacement error (ADE).}
    \label{fig:PS_quantity}
\end{figure}

Figure \ref{fig:PS_quantity} compares the performances (ADE) of patient-specific and multi-patient models, trained on T1 data, and evaluated on both T1 and T2 sessions. As expected, performances on T2 were lower than on T1 for both models due to anatomical and setup variations between planning and treatment. However, the degradation was more pronounced for PS models, indicating reduced robustness across sessions. Despite this drop, PS models consistently outperformed MP models for all training data quantities, on both T1 and T2 evaluations. This difference is statistically significant (paired t-test, $p < 0.05$).

\subsection{Comparison of patient-specific and multi-patient approaches}
Table \ref{table:PS-model_accuracy_centered} compares the performance of both approaches trained with 25,000 T1-DRRs. Evaluation was conducted on both the SM and SV datasets using ADE and FDE as metrics. Overall, the patient-specific (PS) model achieved lower average errors (ADE) than the multi-patient (MP) model. The MP model, however, exhibited consistent accuracy across the entire prediction horizon, on both T1 and T2 images. Notably, when evaluated on T2 data, the performance difference between PS and MP models is no longer statistically significant ($p > 0.05$).
\begin{table}[htpb]
    \centering
    \caption{Comparison of the patient-specific approach and multi-patient on SM and SV datasets: PS models outperform MP models on both T1 and T2 sessions. Performance on the SV dataset is slightly lower due to its higher complexity. Performance is evaluated using the average displacement error (ADE) and the final displacement error (FDE).}
    \label{table:PS-model_accuracy_centered}
    \vspace*{2ex}
    \begin{tabular}{|c|c|c|c|c|c|}
        \hline
        \multirow{2}{*}{\textbf{Dataset}} & \textbf{Training} 
        & \multicolumn{2}{c|}{\textbf{T1 (mm)}} 
        & \multicolumn{2}{c|}{\textbf{T2 (mm)}} \\ 
        \cline{3-6} 
        & \textbf{strategy}  
        & \textit{ADE} & \textit{FDE} 
        & \textit{ADE} & \textit{FDE} \\ \hline
        \multirow{2}{*}{\textit{SM dataset}} 
        & \rule{0pt}{2.5ex}PS & $0.32 \pm 0.31$ & $0.37 \pm 0.36$ & $0.80 \pm 0.33$ & $0.85 \pm 0.39$ \\ 
        & MP & $0.73 \pm 0.50$ & $0.73 \pm 0.51$ & $0.86 \pm 0.37$ & $0.85 \pm 0.36$ \\ \hline
        
        \multirow{2}{*}{\textit{SV dataset}} 
        & \rule{0pt}{2.5ex}PS & $0.60 \pm 0.43$ & $0.76 \pm 0.59$ & $1.36 \pm 1.03$ & $1.48 \pm 1.08$ \\
        & MP & $1.49 \pm 0.89$ & $1.50 \pm 0.92$ & $1.60 \pm 1.03$ & $1.60 \pm 1.04$ \\ \hline

        \multirow{2}{*}{\textit{Averaged}} 
        & \rule{0pt}{2.5ex}PS & $0.48 \pm 0.41$ & $0.59 \pm 0.54$ & $1.11 \pm 0.85$ & $1.21 \pm 0.91$ \\
        & MP & $1.16 \pm 0.84$ & $1.16 \pm 0.86$ & $1.28 \pm 0.89$ & $1.27 \pm 0.90$ \\ \hline
    \end{tabular}
\end{table}
\section{Discussion}
\subsection{Accuracy of the proposed method}
\begin{figure}[htpb]
    \centering    
    \includegraphics[width=\linewidth]{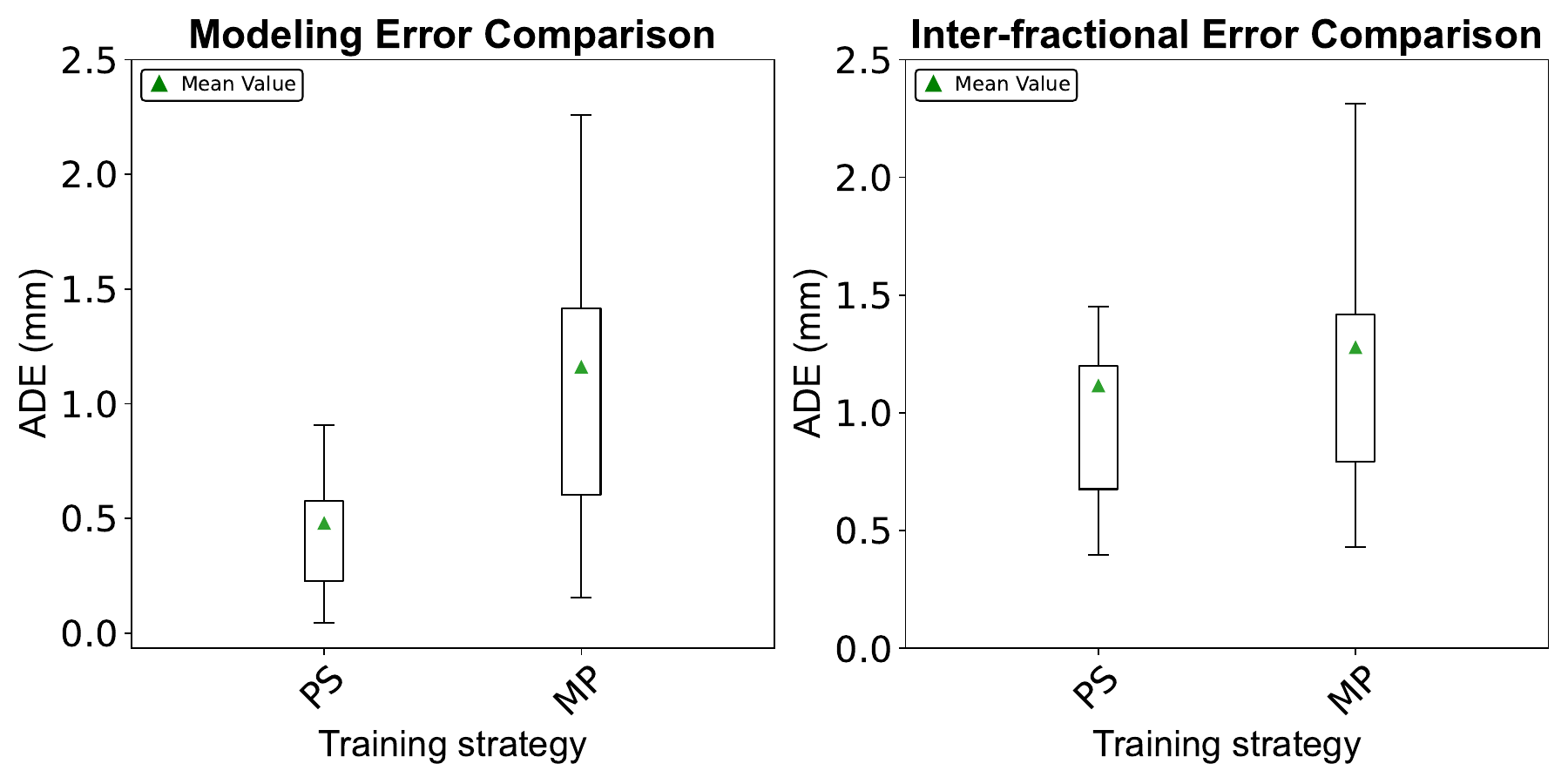}
    \caption{Comparison of modeling and inter-fractional errors between PS and MP strategies across patients. PS training improves modeling error, while both approaches perform similarly in inter-fractional error. Performance is evaluated using the average displacement error (ADE).}
    \label{fig:error}
\end{figure}

The forecasting error in tumor motion prediction is influenced by two key components: modeling error, which reflects the transformer's ability to approximate tumor motion during training, and inter-fractional error, which arises from anatomical variations and setup inconsistencies between treatment sessions. We analyzed these errors under both patient-specific and multi-patient training strategies using 25,000 training samples. Modeling error was assessed on T1 data, while inter-fractional error was evaluated on T2 data. Figure \ref{fig:error} presents the distribution of these errors and compares them with the clinical accuracy achieved with fiducial gold markers ($\pm 2$ mm \cite{imura2005insertion}).

The PS models exhibited significantly lower modeling errors than the MP models, which emphasizes the benefit of tailoring models to the specific motion patterns of individual patients.  This is particularly relevant in radiotherapy, where inter-patient variability can be substantial, making it difficult for a generalized model to capture the unique motion characteristics of each individual. However, from a clinical perspective, inter-fractional error is more critical. As shown in Figure \ref{fig:error}, both strategies yield similar inter-fractional errors, suggesting that most of the residual uncertainty arises from anatomical variations and patient setup differences between treatment sessions. These inter-fractional deviations are largely independent of the modeling approach, as they reflect real physiological and geometric changes rather than algorithmic limitations. Moreover, Table \ref{table:PS-model_accuracy_centered} shows that there is no statistically significant difference in FDE between the two approaches. Hence, the clinical benefit of PS models is limited, as it requires retraining for each patient without reducing inter-fractional error. \\

Additionally, the inter-fractional accuracy of the proposed method exceeded that of implanted fiducial markers ($\pm 2$ mm) in 29 patients with the PS model and in 27 patients with the MP model, highlighting the potential of our markerless approach.
\subsection{Comparison with other markerless methods}
To the best of our knowledge, existing markerless methods focus primarily on tracking rather than forecasting tumor motion. Consequently, we compared our multi-patient approach with a forecasting horizon of $200$ ms. For a fair comparison, we focused on lung tumors and limited the analysis to methods that predict motion in all three clinically relevant directions: superior-inferior, left-right, and anterior-posterior. Reported tracking errors in the literature include $1.90 \pm 0.65$ mm, $1.03 \pm 0.34$ mm, and $1.07 \pm 0.35$ mm for studies with small patient cohorts (5, 8, and 10 patients, respectively)\cite{hirai2019real,sakata2020machine,terunuma2023explainability}. In contrast, our approach, evaluated on a larger cohort of 32 patients, achieved an accuracy of $1.03 \pm 0.77$ mm, yielding results comparable to those reported in the literature.
\subsection{Limitations and further works}
\subsubsection{Fluoroscopy images}
The current approach was validated on digitally reconstructed radiographs (DRRs) rather than on real fluoroscopy images, which introduces a potential domain gap. This discrepancy may affect generalizability in clinical settings, particularly due to differences in noise, contrast, and anatomical details. Additionally, both approaches were trained using low-resolution images ($64 \times 64$), which may limit their ability to capture fine anatomical details. While this constraint affects both approaches, it could have a more pronounced impact on patient-specific models, which rely on learning from data that is less diverse in terms of anatomy and imaging variability. As a result, the limited spatial resolution might underestimate the potential of patient-specific training compared with a multi-patient setup. Further investigation is needed to evaluate how increasing resolution affects each approach, while respecting the clinical time constraint between planning and treatment sessions.

\subsubsection{Fine-tuning}
From a clinical standpoint, our results indicate that while patient-specific models achieve lower modeling errors on planning data, their lack of robustness to inter-fractional anatomical changes limits their clinical utility. In contrast, multi-patient models trained offline generalize well across patients and maintain stable performance on treatment data, making them a promising non-invasive alternative to fiducial marker-based tracking systems. \\

To bridge the gap between accuracy and generalization, we propose a hybrid strategy: an MP model can be pretrained offline on a large cohort and subsequently fine-tuned on  a new patient's data within a few minutes. Such a strategy has been shown~\cite{sbad,manjahholonic} to combine structured and global feature learning from the general-purpose model while tailoring the model to individual data distributions. This hybrid approach could support real-time tumor tracking in clinical workflows while minimizing patient burden and system complexity, all without the need for invasive marker implantation. Specifically, the proposed method could provide tighter dose conformality while sparing healthy tissues.

\section{Conclusions}
This study is the first to explore vision transformers for markerless lung tumor motion forecasting in proton therapy. We compared two training strategies — patient-specific (PS)  and multi-patient (MP) — under realistic clinical constraints that reflect the limited time and data available between planning and treatment. The patient-specific approach prioritizes individualized features but is constrained by the limited training time available between the planning and treatment sessions. In contrast, the multi-patient approach captures a broader distribution of data and is free from these timing constraints.

Our results show that PS models achieve higher accuracy on planning data, but their performance deteriorates when applied to treatment data, limiting their clinical applicability. In contrast, MP models trained offline on diverse patient cohorts offer robust, stable performance across sessions, demonstrating their potential as a non-invasive alternative to fiducial marker-based tracking.

To combine the strengths of both approaches, we propose a clinically viable hybrid strategy: an MP model pretrained offline can be rapidly fine-tuned on a new patient's treatment data within minutes. This enables precise tumor motion forecasting tailored to the individual patient, while preserving the generalization and deployment advantages of the MP model. Such a solution could support reliable, markerless, real-time tracking in adaptive radiotherapy workflows without increasing patient burden.


\section{Acknowledgments}
Gauthier Rotsart de Hertaing and Dani Manjah were supported by the Walloon region under grant n°2010149 - ARIES. We also thank the OpenHub team of UCLouvain for providing computational resources. 
\bibliographystyle{unsrtnat}
\bibliography{references}  






\end{document}